\documentclass{article}

\usepackage[preprint]{neurips_2019}

\usepackage[utf8]{inputenc} 
\usepackage[T1]{fontenc}    
\usepackage{hyperref}       
\usepackage{url}            
\usepackage{booktabs}       
\usepackage{amsfonts}       
\usepackage{nicefrac}       
\usepackage{microtype}      

\usepackage{amsmath}
\usepackage{graphicx}
\usepackage{subfloat}
\usepackage{xcolor}
\usepackage{wrapfig}
\usepackage{subfig}

\title{MolecularRNN: Generating realistic molecular graphs with optimized properties}

\author{%
  Mariya Popova \\
  University of North Carolina at Chapel Hill\\
  Curriculum in Bioinformatics and\\
  Computational Biology \\
  Chapel Hill, NC 27599 \\
  \texttt{mariya@live.unc.edu} \\
  \And
  Mykhailo Shvets \\
  University of North Carolina at Chapel Hill\\
  Department of Computer Science \\
  Chapel Hill, NC 27599 \\
  \texttt{mshvets@cs.unc.edu} \\
  \And
  Junier Oliva \\
  University of North Carolina at Chapel Hill\\
  Department of Computer Science \\
  Chapel Hill, NC 27599 \\
  \texttt{joliva@cs.unc.edu} \\
  \And
  Olexandr Isayev \\
  University of North Carolina at Chapel Hill\\
  Eshelman School of Pharmacy \\
  Chapel Hill, NC 27599 \\
  \texttt{olexandr@olexandrisayev.com} \\
  \And
}

\begin{document}

\maketitle

\begin{abstract}
 Designing new molecules with a set of predefined properties is a core problem in modern drug discovery and development. There is a growing need for {\it de-novo} design methods that would address this problem. We present MolecularRNN, the graph recurrent generative model for molecular structures. Our model generates diverse realistic molecular graphs after likelihood pretraining on a big database of molecules. We perform an analysis of our pretrained models on large-scale generated datasets of 1 million samples. Further, the model is tuned with policy gradient algorithm, provided a critic that estimates the reward for the property of interest. We show a significant distribution shift to the desired range for lipophilicity, drug-likeness, and melting point outperforming state-of-the-art works. With the use of rejection sampling based on valency constraints, our model yields 100\% validity. Moreover, we show that invalid molecules provide a rich signal to the model through the use of structure penalty in our reinforcement learning pipeline.
 
\end{abstract}

\section{Introduction}

The process of discovering a new drug candidate, passing it through clinical trials and onto the market is extremely hard, time-consuming, and expensive. Less than one out of every 10,000 drug candidates becomes an approved marketed drug. Only three out of every 20 approved drugs bring in enough revenue to cover developmental costs. Moreover, it takes approximately 10-15 years and the average cost of \$1-3 billion to develop each new drug. The development of computer algorithms can help in this process, for example, by suggesting novel molecules with optimal property profiles. This process is called \textit{de novo} molecular design. The goal of \textit{de novo} methods is to create novel molecules with desired properties. It typically comprises from three tasks: 1) molecule generation; 2) scoring, and 3) optimization \citep{schneider2005computer}. Each of these steps could be performed sequentially or together by either human expert or machine.

Machine learning systems are radically transforming the practice of chemical and molecular sciences \citep{butler2018ml}. Drug discovery is well positioned to be the next frontier for a potential breakthrough. Not surprisingly, recent advances in machine learning methods have also facilitated the automated generation of new molecules with the desired properties. Recently we have seen a huge rise due to deep neural networks, which are now well-developed and optimized for continuous signals with naturally defined neighborhoods for the simplest elements. For example, pixels are grouped in local neighborhoods in images with convolutional neural networks, and words are grouped in natural language with recurrent neural networks. However, graphs are a more complex structure with non-uniform neighborhoods, for which networks have been introduced not so long ago.

While there exist several representations of molecules (SMILES, fingerprints, 3D atom configuration), graphs are the most natural one, with direct mapping of atoms into nodes, and bonds into edges. A molecular graph is undirected but has several node types (carbon, oxygen, nitrogen, etc.), as well as  edge types (single, double, triple, and aromatic bond). Other representations suffer from being complicated and lacking a clear notion of distance. For example, SMILES (natural language representation of a molecule in the form of a string), adds a layer of complexity with its grammatical rules, but more importantly, a pair of molecules which share a common scaffold (core), can be encoded by very different SMILES strings representations. We introduce a molecular graph recurrent generative model, showing that incremental molecular graph construction seamlessly incorporates the proposed valency-based rejection sampling procedure that yields 100\% valid molecules during inference, while also getting signal from invalid intermediate molecules through our structural penalty.

After training our model in an unsupervised manner to match the distribution of large training datasets, optimizing properties is particularly interesting for application. We show the capability of optimizing the generated molecules to a specific property range through reinforcement learning, where the reward is constructed based on the output of a critic.

We summarize our main contributions as the following:

\begin{itemize}
    \item molecular graph recurrent model, MolecularRNN, for direct generation of realistic molecular graph structures that shows high validity/uniqueness/novelty
    \item valency-based rejection sampling method during inference that produces 100\% valid molecules, and the structural penalty during training for atoms violating valency constraints
    \item target property optimization with reinforcement learning for improving drug-likeness, lipophilicity, and melting temperature
    \item an unprecedented large-scale experimental analysis and application through amalgamation of these techniques.
\end{itemize}

\section{Related work}

Various approaches to computational {\it de-novo} molecule generation have been proposed. The fundamental differences in these approaches lie in types of molecules representation.  The most well-studied way to represent a chemical molecule is a simplified molecular-input line-entry system (SMILES) string \citep{weininger1988}. SMILES string consists of symbols corresponding to nodes of the molecular graph in their depth-first order, unambiguously describing the composition and structure of the chemical molecule.
Approaches for generating molecules as SMILES strings are using a recurrent neural network to learn a language model of SMILES \citep{olivecrona2017molecular,GomezBombarelli, popova2018deep}. Probably the biggest limitation of these methods is imperfect validity (i.e. some of the generated samples are not chemically valid molecules) due to a challenge of learning complex grammatical rules. Another limitation is that SMILES-based approaches cannot be naturally extended to scaffold optimization when a generation process starts from a given core of the molecule and the task is to find a molecule with better properties and pattern of substituents while maintaining the same molecular core.  

Another way to represent a chemical molecule is through its molecular graph. 
Graph-based approaches typically do not suffer from the problem of invalidity of generated molecules. It is also possible to enforce physical constraints on the valency, i.e., how many neighbors each atom can have depending on the atom type. Moreover, these models are more interpretable and more intuitive to chemists. Various algorithms for generating molecular graphs have been developed \citep{jin2018junctiontree, li2018learning}. 
\citet{jin2018junctiontree} proposed a junction tree variational autoencoder. This model first generates a junction tree where every node corresponds to a structural fragment rather than a single atom. Then, the junction tree is converted into a valid molecule with a sampling procedure. This approach produces valid molecules by design; however, there is ambiguity in the process of converting a generated junction tree into a molecule due to sampling. While this is not a problem for the unconstrained generative process, it may cause difficulties with property optimization because molecules with the same junction tree may have a drastic difference in property value. \cite{jin2018junctiontree} argue that it is beneficial to generate a graph from fragments, however, atom-by-atom models have already proven as a strong baseline \citep{you2018graphconvpolicy, liu2018constrainedgraph, li2018multi}.
In \citep{li2018learning} the process of graph generation is sequential. Nodes are generated one at a time and then connected to the existing partial graph. With a sequential process, the same graph can be generated with multiple sequences of steps due to the node order permutation. This work does not address the problem of node order permutation. Another limitation of this work is the constraints on the graph size. Only molecular graphs with at most $20$ heavy atoms were considered which is not enough for any practical purpose.

In \citep{you2018graphconvpolicy}, the procedure of molecular graph generation is presented as a Markov Decision Process.  The model uses graph convolutional network (GCN) model for goal-directed graph generation with reinforcement learning and adversarial training. This work similar to \citep{jin2018junctiontree} only reports top 3 molecules, while top 3 may not represent the model performance as well as the distribution of a property obtained from a large number of generated samples. Recently, GraphRNN model~\citep{you2018graphrnn} was proposed for the generation of undirected graphs. We extend this model to include node and edge types predictions.  

Previous works have explored a variety of methods to optimize properties of interest for molecules: fine-tuning \citep{olivecrona2017molecular}, transfer learning \citep{segler2017generating}, reinforcement learning \citep{popova2018deep} and adversarial training \citep{kadurin2017drugan}. Often physicochemical properties, such as the octanol-water partition coefficient (logP) and molecular weight (MW), melting temperature ($T_{melt}$) are used as a convenient proxy for drug-likeness and chances of a particular molecule for the successful drug candidate. 


\section{Methods}

The core of our approach is a MolecularRNN model, which extends GraphRNN~\citep{you2018graphrnn} model for generating graphs with node and edge types. Section~\ref{sec:method:graphrnn} gives background on GraphRNN, and the extension is described in Section~\ref{sec:method:moleculernn}. We introduce a method of valency-based rejection sampling in Section~\ref{sec:method:valency_check} that yields 100\% validity in inference mode. We show a distribution shift towards desired properties values with reinforcement learning in Section~\ref{sec:rl}. Finally in Section~\ref{sec:str_rew} we introduce our structural penalty that provides a signal from the invalid samples during training.


\subsection{Background: GraphRNN model}
\label{sec:method:graphrnn}

GraphRNN \citep{you2018graphrnn} was introduced for generation of undirected graphs $G=(V, E)$ with a set of $n$ nodes $V=\left(v_1, \dots, v_n\right)$ and a set of undirected edges $E=\left(\{v_i, v_j\} | v_i, v_j \in V\right)$ between those nodes. Under some node ordering $\pi$ this graph is represented with its adjacency matrix $A^\pi \in \{0, 1\}^{n \times n}$ with $A^\pi_{i,j} = 1 \text{ iff } \left(\pi(v_i), \pi(v_j)\right) \in E$. The model generates graphs as sequences of adjacency vectors $S^\pi_i \in \{0, 1\}^{i-1}$ from node $\pi(v_i)$ to previous nodes under $\pi$. Thus, 
$S^\pi_i = \left(A_{1, i}^\pi, \dots A_{i-1, i}^\pi \right)^T$, and likelihood $p(S^{\pi})$ can be modelled sequentially, being decomposed as
\begin{equation}
 p(S^{\pi}) = \prod\limits_{i=1}^{n+1} p(S_i^\pi | S_1^\pi, \dots, S_{i-1}^\pi) = \prod\limits_{i=1}^{n+1}p(S_i^\pi | S_{<i}^{\pi})
 \label{eq:graphrnn_likelihood}
\end{equation}
with the special end of sequence token (EOS) as an extra node $n+1$.

State-transition function carries the information from step $i-1$ to step $i$, generating a node, and output function predicts the parameters for sampling current adjacency vector $S_i^\pi$ of edges. According to GraphRNN, we consider recurrent neural networks for both  state-transition function ({\it NodeRNN}) and output function ({\it EdgeRNN}). Thus, NodeRNN unrolls across nodes, updating its hidden state, while EdgeRNN unrolls across edges from $i$ to previous nodes, creating parameters $\theta_{i,j}$ with the use of a small MLP head with sigmoid activation, which models $S_i^{\pi}$ as a dependent Bernoulli sequence:

\begin{equation}
\begin{aligned}
    & h^{node}_i = \text{NodeRNN}(h_{i-1}^{node}, S_{i-1}^\pi), & h^{node}_0 &= \mathbf{0} \\
    & h^{edge}_{i,j} = \text{EdgeRNN}(h_{i,j-1}^{edge}, S_{i,j-1}^\pi), & h^{edge}_{i,0} &= h_i^{node}\\
    & \theta_{i,j} = \text{EdgeMLP}(h^{edge}_{i,j}) \\
\end{aligned}
\label{eq:graphrnn_formulas}
\end{equation}

One of the key insights of the method is to re-order the nodes with breadth-first search (BFS), starting from $\pi(v_1)$, which gradually reduces the space complexity for graph representations. Moreover, BFS order also reduces the number of edge predictions that have to be made, limiting the size of $S_i^\pi$ to $M$ dimensions, which appears to be a small number in practical tasks. Thus, for our modified MolecularRNN (Section~\ref{sec:method:moleculernn}) we empirically establish $M=12$.

\subsection{MolecularRNN}
\label{sec:method:moleculernn}

In order to represent a molecule with a graph, atoms are mapped to nodes, while bonds are mapped to edges. Now, adjacency vector entries represent categorical bond types $S_{i,j}^{\pi} \in \{0, 1, 2, 3\}$, corresponding to no, single, double, and triple bonds (molecules are modeled in kekulized form as defined in RDKit~\citep{landrum2006rdkit}). Similarly, categorical type $C_i^\pi \in \{1, 2, \dots, K\}$ (oxygen, nitrogen, chlorine, etc.) is assigned to each node. Notice that here a node always has a valid atom class. That is, there is no "terminal node" class, as terminal node notion is already incorporated into $S_i^\pi$. Specifically, when a node is generated that has no edges to any of the previous nodes, such a node is terminal. Atom class prediction is ignored for this node in our setting.

Likelihood in Equation~\ref{eq:graphrnn_likelihood} is rewritten accordingly for MolecularRNN:

\begin{equation}
    p(S^\pi, C^\pi) = \prod\limits_{i=1}^{n+1}p(C_i^\pi | S_{<i}^\pi, C_{<i}^\pi) p(S_i^\pi | C_i^\pi, S_{<i}^\pi, C_{<i}^\pi),
\label{eq:moleculernn_likelihood}
\end{equation}

with $p(C_{n+1} | S_{<n+1}, C_{<n+1}) \equiv 1$ for the terminal node $n+1$.

\begin{figure}[ht]
 \includegraphics[width=\textwidth]{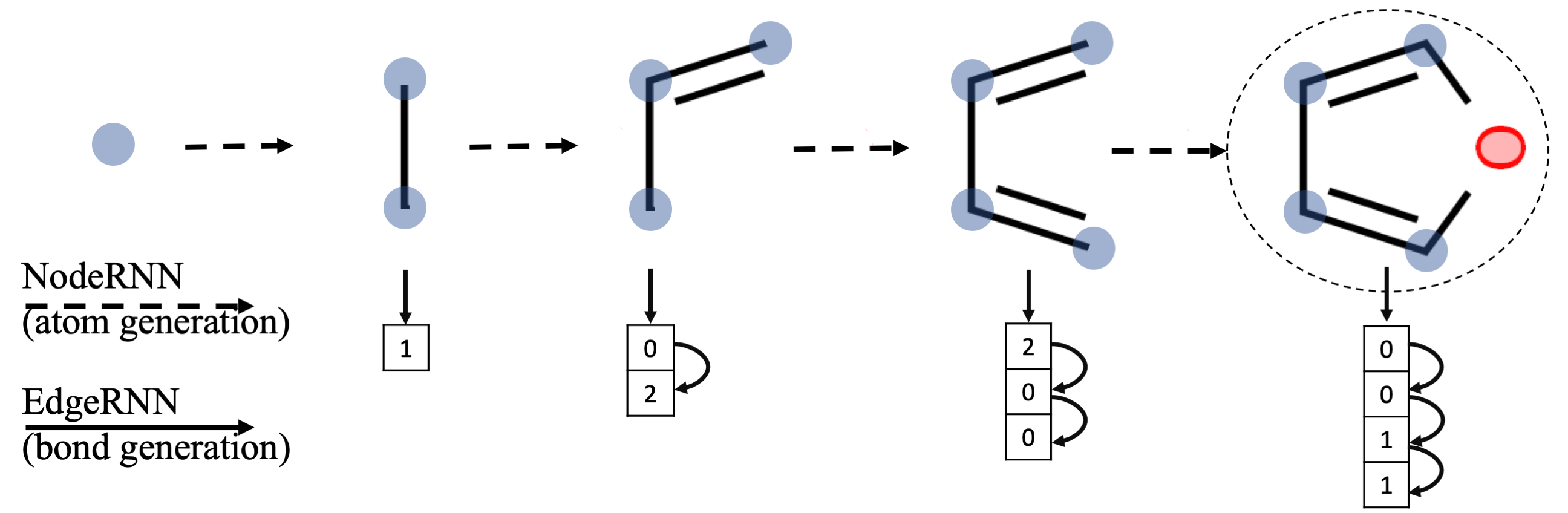}
 \caption{MolecularRNN model. The model consists of NodeRNN that unrolls across atoms, predicting the type of the next atom in the molecular graph, and EdgeRNN that for every atom is initialized with NodeRNN hidden state, and unrolls across preceding atoms to predict bond types.}
 \label{fig:moleculernn}
\end{figure}

In our model, once the sub-graph on the first $i-1$ nodes under permutation $\pi$ is completed, NodeRNN can momentarily decide on the atom type of the following node $i$. Thus, the process represents a dependent multivariate distribution. Accounting for the sub-graph, as well as the $i$-th atom type, the model switches to EdgeRNN that links the newly generated node to the set $\{1, \dots, i-1\}$. That step is in turn modeled with a dependent multivariate distribution, as EdgeRNN is unrolled across nodes that precede $i$. Overall MolecularRNN structure is shown in Figure~\ref{fig:moleculernn}. The model uses embeddings for categorical inputs, and a two-layer MLP with softmax output activation is added on top of hidden states $h^{node}$ and $h^{edge}$ for categorical prediction, so Equation~\ref{eq:graphrnn_formulas} is modified:

\begin{equation}
\begin{aligned}
    & \text{input}_{i-1} = [emb(S_{i-1}^{\pi}), emb(C_{i-1}^{\pi})] \\
    & h_i^{node} = \text{NodeRNN}(h_{i-1}^{node}, \text{input}_{i-1}), & h_0^{node} &= \mathbf{0} \\
    & \psi_i = \text{NodeMLP}(h_i^{node}) \\
    & h^{edge}_{i,j} = \text{EdgeRNN}(h_{i,j-1}^{edge}, emb(S_{i,j-1}^\pi)), & h^{edge}_{i,0} &= h_i^{node}\\
    & \phi_{i,j} =\text{EdgeMLP}(h^{edge}_{i,j})    ,
\end{aligned}
\end{equation}

In our BFS ordering the first node is always a Carbon atom, since every organic molecule contains at least one such atom.

\subsection{Valency-based rejection sampling}
\label{sec:method:valency_check}

As we have seen, MolecularRNN samples edge types on each sub-step from a multinomial distribution with parameters coming out of softmax predictions. Even when the model is trained well for producing valid molecules, the softmax layer prediction will always have nonzero values, so if sampling is arbitrarily long, any graph can be sampled from the support space. However, real molecules have valency constraints. That is, per-atom valency has to be respected to satisfy chemical constraints. Consequently, in each step, we can ensure that the current sum of all bonds does not exceed the allowed valency. When generating an edge corresponding to a bond of order $k$ between $i$ and $j$ we check the rejection sampling constraint for both atoms:

\begin{equation}
\sum\limits_{j}A_{i,j}^\pi + k \leq \mathrm{valency}_{C^{\pi}_{i}} \quad \text{and} \quad
\sum\limits_{i}A_{i,j}^\pi + k \leq \mathrm{valency}_{C^{\pi}_{j}} 
\end{equation}

For the final molecule, atoms that have not filled up their valencies are complemented with Hydrogens. Notice that valency can be directly enforced only for graphs, unlike SMILES representation, where intermediate sub-strings are not chemically meaningful.

\subsection{Property optimization}
\label{sec:rl}

While generating realistic molecules is an appealing goal, our ultimate aim is to shift the distribution of the generated samples for some desired property. To optimize the chosen property, we use policy gradient algorithm. In this formulation, MolecularRNN acts as a policy network which outputs probability of the next action given the current state. The set of actions is defined as the set of atom labels times the set of combinations of possible generated atom connection to the existing graph. The set of states is defined as all possible sub-graphs of graphs with up to a fixed number of $N$ nodes. Consistently with the BFS ordering in MolecularRNN, initial state $s_0$ is a graph of a single carbon atom. The set of final states is defined as the set of all graphs that correspond to a valid molecule with up to $N$ heavy atoms. The reward $r(s_N)$ for a final state $s_N$ (without loss of generality $s_N$ is used even if $n<N$ in the generated graph) is calculated with a critic. We distributed the final reward to all intermediate steps, with the discounting factor, which proves to show more stable convergence in our experiments. Thus, intermediate rewards $r(s_i), ~0<i<N$ are obtained by discounting the final reward with a fixed factor $\gamma$. 

The transition probabilities $p(s_i|s_{i-1}; \theta)$ are the elements of the product in Equation~\ref{eq:moleculernn_likelihood}.
Given those, we can write down the loss function for the policy gradient optimization algorithm by \citet{williams1987}, which is designed to maximize the expected reward:
\begin{equation}
    \label{eq:objective}
    L(\theta) = -\sum_{i=1}^N r(s_N) \cdot \gamma^{i} \cdot \log{p\left( s_i | s_{i-1}; \theta\right)}.
\end{equation}

\subsection{Structural penalty}
\label{sec:str_rew}
Valency-based rejection sampling can be used in inference, as was already described. However, the invalid intermediate structures that are obtained during training can provide a useful signal to the model. For example, a molecule can be {\it almost realistic} except for few invalid bonds. We introduce an additional structure penalty for the atoms that disrespect valencies. Thus, instead of providing a penalty for the whole molecule, we target specific atoms, which results in the modification of parameters that respect valency constraints.

\section{Experiments}


To validate the quality of our results and compare those to the state-of-the-art methods we use validity, uniqueness, novelty, internal diversity, synthetic accessibility score (SA score)~\citep{ertl2009estimation} and drug-likeness score (QED)~\citep{bickerton2012quantifying}. Validity is the percentage of chemically valid molecules.
Uniqueness is the percentage of unique molecules in the generated pool. Notice that uniqueness is highly dependent on the pool size, and may significantly drop for a large generated library. In our experiments, we report uniqueness in up to a million of generated samples. Internal diversity, as proposed in MOSES benchmark \citep{polykovskiy2018molecular}, is a quantitative metric of the richness of the generated library and is calculated as the average pairwise distance between all pairs of molecules in the library. 
SA score is an estimation of how hard is to synthesize a given molecule, which also reflects its structural complexity. Molecules with a higher score will be more complex and harder to synthesize. However, molecules with very low score might be not complex enough to have the desired property. The values of interest for this metric are in the range between 2 and 4. Finally, QED is a measure of drug-likeness in the range from $0$ to $1$. 

\subsection{Unsupervised likelihood training}




We first pretrain MolecularRNN on a large unlabeled dataset of molecules to teach the model to generate diverse realistic samples. Three training datasets are used: ChEMBL~\citep{gaulton2011}, random 250k molecules from ZINC~\citep{irwin2005zinc} and MOSES~\citep{polykovskiy2018molecular}. These three datasets have different statistics. The statistics are shown in Table~\ref{tab:data_stats}. ChEMBL dataset contains around $1.5$ million of real bioactive molecules (every molecule has at least one experimental bioactivity measurement) and is the most diverse out of all three datasets that we considered. ZINC 250k random dataset contains 250 thousand molecules randomly selected from a database of commercially available compounds \citep{irwin2005zinc}. MOSES dataset contains almost 2 million molecules that were selected from the ZINC database based on several filters to only include molecules with drug-like properties.

\begin{table}[h!]
    \caption{Statistics for training datasets.}
    \centering
    \begin{tabular}{l|ccc}
    \toprule
        & \textbf{ChEMBL} & \textbf{ZINC} & \textbf{MOSES} \\
        \midrule
        Number of molecules & $1507869$ & $249456$ & $1936962$\\
        Mean molecular weight & $389 \pm 103$ & $331 \pm 62$ & $307 \pm 28$ \\
        QED score & $0.56 \pm 0.21$ & $0.73 \pm 0.14$ & $0.81 \pm 0.09$ \\
        SA score & $2.88 \pm 0.80$ & $3.05 \pm 0.83$ & $2.45 \pm 0.46$ \\
        \bottomrule
    \end{tabular}
    \label{tab:data_stats}
\end{table}

We considered $9$ most common elements ($C, N, O, F, P, S, Cl, Br, I$) and $3$ bond types (single, double and triple). The number of atoms in the molecule is restricted to be from $10$ to $50$, which is chosen based on ChEMBL dataset, where 96\% of molecules lie in this range. EdgeRNN is unrolled (as discussed in Section~\ref{sec:method:graphrnn}) for $M=12$ steps for each atom. The following architectural parameters are used in all our settings: node embedding of size $128$, edge embedding of size $16$, NodeRNN with $4$ GRU layers of hidden size $256$ each, $2$ layer NodeMLP with $128$ hidden size and ReLU nonlinearity after the first layer, and EdgeRNN with $4$ GRU layers of hidden size $128$ each. During the unsupervised phase, models are trained with Adam optimizer for $250$ epochs on $4$ GPUs with a per-GPU batch size of $512$. The starting learning rate is $0.001$ with a multiplicative drop of $0.999$ every $k$ iterations, and $k$ is chosen based on the dataset so that the learning rate drops to $10^{-5}$ by the end of the training. MolecularRNN trained with the likelihood maximization on the training datasets achieves validity rate of 65\% without valency-based rejection sampling. We further used structural penalty described in section \ref{sec:str_rew} to shift the model towards generating molecules that respect valency constraints. To that end, every atom that violates its valency constraints is assigned a penalty of $-10$, and then the model is optimized with the policy gradient method. After training with structural penalty, our model achieved valid rate of 90\% without valency-based rejection sampling. Enabling valency-based rejection sampling results in 100\% valid rate for all models.


\begin{table}[h!]
    \caption{Statistics for 1 million molecules generated by 3 models pretrained on 3 training datasets}
    \centering
    \begin{tabular}{l|ccccccc}
    \toprule
    Training set& \textbf{Valid} & \textbf{Unique} & \textbf{Novel} & 
    \begin{tabular}{@{}c@{}}
    \textbf{IntDiv} \\
    \textbf{(p=1)}
    \end{tabular} & 
        \begin{tabular}{@{}c@{}}
    \textbf{IntDiv} \\
    \textbf{(p=2)}
    \end{tabular} & 
    \textbf{SA score} & \textbf{QED} \\
        \midrule
        ChEMBL & 100 \% & 99.2\% & 99.3 \% & 0.895 & 0.890 & $3.67 \pm 1.20$ & $0.56 \pm 0.20$ \\
        ZINC 250k & 100 \% & 99.8 \% & 100 \% & 0.892 & 0.887 & $3.60 \pm 1.01$ & $0.68 \pm 0.16$ \\
        MOSES & 100 \% & 99.4 \% & 100 \% & 0.881 & 0.876 & $3.24 \pm 0.97$ & $0.74 \pm 0.14$ \\
        \bottomrule
    \end{tabular}
    \label{tab:pretrain}
\end{table}

Table~\ref{tab:pretrain} summarizes the results of unsupervised likelihood training of MolecularRNN on the three datasets. Statistics are calculated on $1$ million generated graphs, which is a much larger scale than previously reported. For comparison, \citet{jin2018junctiontree} sample $5$ thousand graphs, and \citet{li2018learning} evaluate a $100$ thousand set. In all cases, the model produces novel diverse realistic molecules. 

We also compare our model with GCPN \citep{you2018graphconvpolicy} and JT-VAE \citep{jin2018junctiontree} in Table~\ref{tab:pretraining_results} on 30K molecules generated from each method. MolecularRNN produces comparable results to the baselines in terms of validity, uniqueness, and novelty. GCPN tends to generate overly complex, hard to synthesize molecules (high SA score). Samples from our model are more realistic, and also have higher internal diversity than the ones from JT-VAE.


\begin{table}[h!]
    \centering
    \caption{Comparison of MolecularRNN, GCPN \citep{you2018graphconvpolicy} and JT-VAE \citep{jin2018junctiontree}. Models are trained on ZINC 250k dataset. Statistics are calculated for 30000 generated molecules.}
    \begin{tabular}{l|cccccc}
        \toprule
        & \textbf{Valid} & \textbf{Unique} & \textbf{Novel} &\textbf{SA score} & \textbf{QED} & \textbf{InvDiv}\\
         \midrule
        JT-VAE \citep{jin2018junctiontree}  &99.8\%
        & 100 \%& 100\%& 3.37 & 0.76 & 0.85 \\
        GCPN \citep{you2018graphconvpolicy}& 
         100\% & 99.97\% & 100\% & 4.62 & 0.61 & $0.90$ \\
        MolecularRNN & 100\% & 99.89\% & 100\% & 3.59 & 0.68 & 0.89\\
        \bottomrule
    \end{tabular}
    \label{tab:pretraining_results}
\end{table}



\subsection{Property optimization with reinforcement learning}
We performed experiments on the properties optimization of generated molecules starting with our strong pretrained model with the policy gradient algorithm (section ~\ref{sec:rl}). We choose maximization of penalized logP as defined in \citep{jin2018junctiontree} and QED \citep{bickerton2012quantifying} starting from MolecularRNN that is likelihood-pretrained on ZINC 250k dataset. We also performed an additional experiment with maximization of melting temperature. Such an analysis has never been reported in graph-based generative models before. This is an appealing exercise because it requires training an additional model for melting temperature prediction, while logP and QED can be computed directly from the molecular graph structure. This experiment mimics realistic drug discovery scenario, where toxicity or bioactivity is optimized. It paves the way for further research in this important direction.

\begin{figure*}[h!]
\centering
\subfloat[penalized logP]{\includegraphics[width=0.45\linewidth]{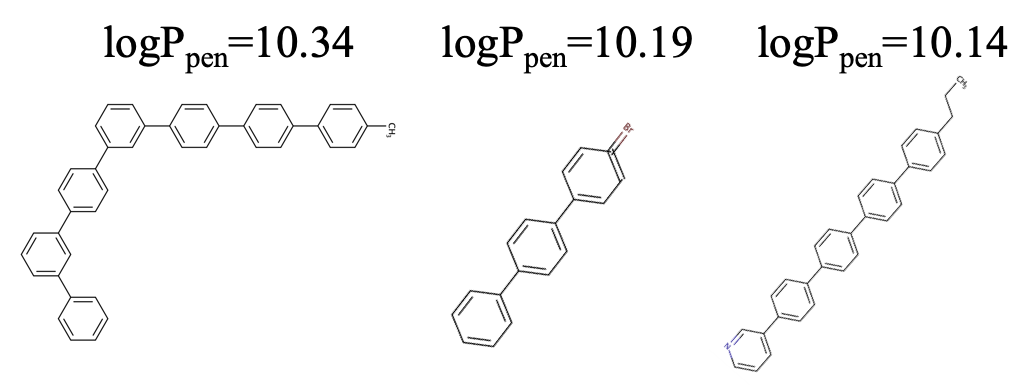}\label{fig:top3_logp}}
\qquad
\subfloat[QED]{\includegraphics[width=0.45\linewidth]{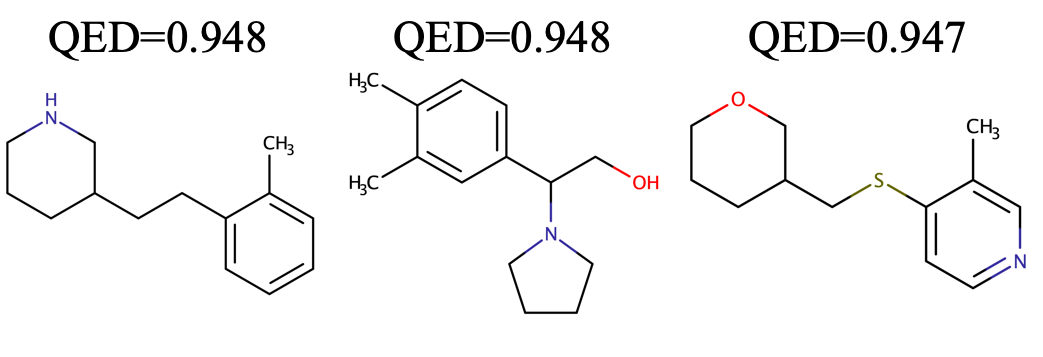}\label{fig:top3_qed}}
\caption{Top 3 molecules for MolecularRNN optimized with policy gradient}
\label{fig:top3_mols}
\end{figure*}

\paragraph{Penalized logP and QED maximization.} As in \citep{you2018graphconvpolicy, jin2018junctiontree}, we independently maximize two properties -- penalized logP and QED score. MolecularRNN is tuned for 300 iterations with a generated batch size of 512 and Adam optimizer with a constant learning rate of $10^{-5}$. The objective function in Equation~\ref{eq:objective} maximizes the following rewards:
\begin{equation*}
    r(mol) = 5\cdot \log{P}_{pen}(mol) \\ 
\end{equation*}
\begin{equation*}
    r(mol) = 10\cdot QED(mol).
\end{equation*}
We use discount factor $\gamma=0.97$. The best 3 molecules after optimization for both properties are shown in Table \ref{tab:top3_results}, and  demonstrates the distribution shift. In this experiment, our model outperforms all baselines in both tasks. The top 3 molecules are shown in Figure \ref{fig:top3_mols}. Samples with high logP values are very realistic, as the model learned to grow a chain of aromatic rings that would very strongly bind to a lipid membrane (high lipophilicity). This is an indicator that the model learned some underlying physics about relationship between molecular structure and properties.

\begin{table}[h!]
    \caption{Comparison of the top 3 scores for penalized logP and QED.}
    \centering
    \begin{tabular}{l|cccc|cccc}
        \toprule
         & \multicolumn{4}{c}{Penalized logP} & \multicolumn{4}{c}{QED score} \\
         \midrule
        \bf{Methods} & \bf{1st} & \bf{2nd} & \bf{3rd} & \bf{Valid} & \bf{1st} & \bf{2nd} & \bf{3rd} & \bf{Valid}\\
        \midrule
        ZINC & 4.52 & 4.30 & 4.23 & 100\%& 0.948 & 0.948 & 0.948 & 100\%\\
        \citep{guimaraes2017objective}    &
        3.63 & 3.49 & 3.44 & 0.4\% & 0.896 & 0.824 & 0.820 & 2.2\% \\
        JT-VAE \citep{jin2018junctiontree} &
        5.30 & 4.93 & 4.49 & 100\%  & 0.925 & 0.911 & 0.910 & 100\%  \\
        GCPN \citep{you2018graphconvpolicy} & 
         7.98 & 7.85 & 7.80 & 100\%  & 0.948 & 0.947 & 0.946 & 100\% \\
        MolecularRNN & \bf{10.34} & \bf{10.19} & \bf{10.14} & \bf{100\%} & \bf{0.948} & \bf{0.948} & \bf{0.947} & \bf{100\%}\\
        \bottomrule
    \end{tabular}
    \label{tab:top3_results}
\end{table}


\begin{wrapfigure}[11]{r}{0.4\linewidth} 
\vspace{-1ex}
\includegraphics[width=\linewidth]{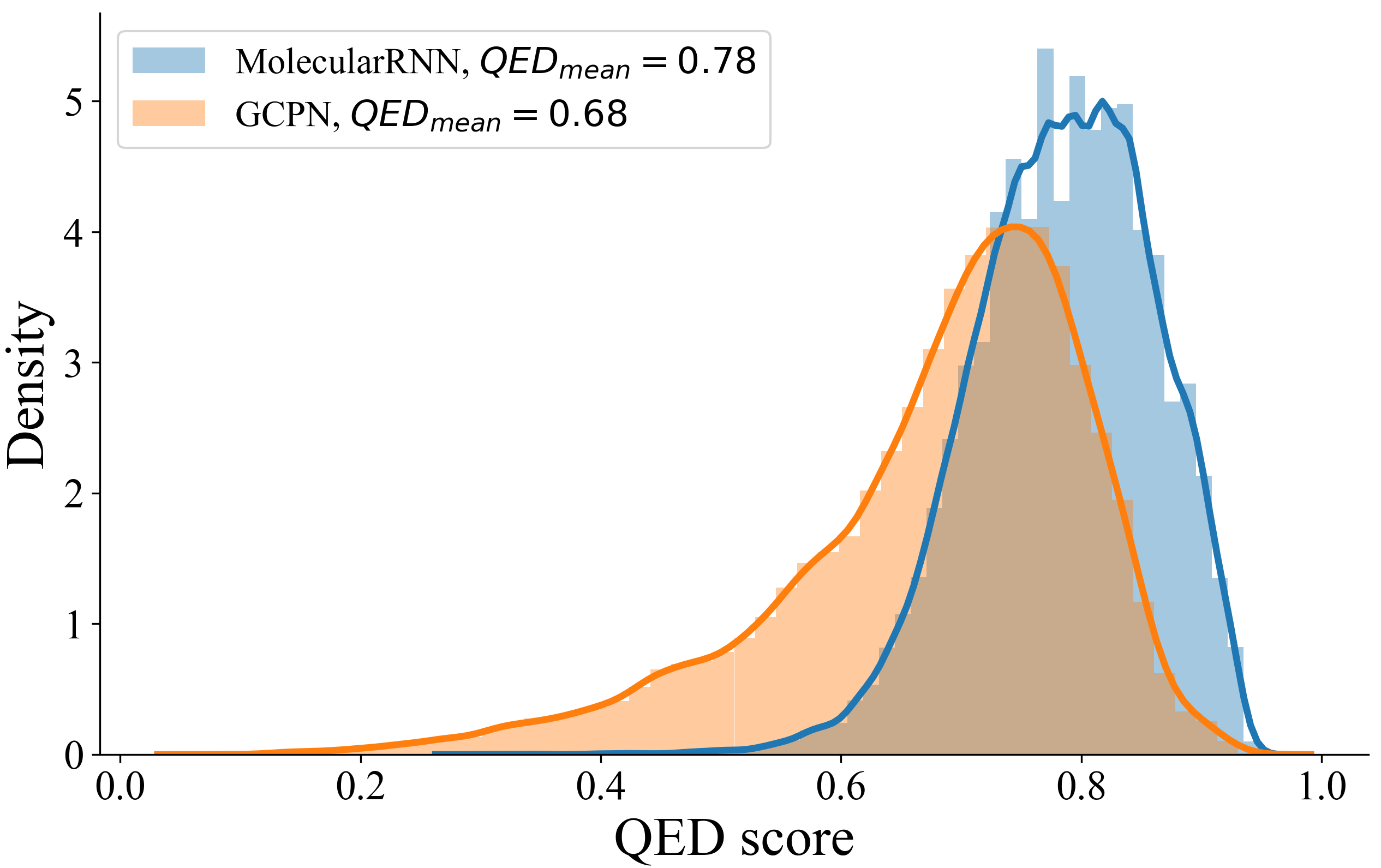}
\caption{Distribution of maximized QED for MolecularRNN and GCPN.}
\label{fig:qed}
\end{wrapfigure}

We took a step further and not only looked at molecules with top 3 scores but also considered the full distribution of the maximized QED for libraries generated with our MolecularRNN and GCPN \citet{you2018graphconvpolicy} as the best baseline. We argue that reporting only the top 3 scores is not the most informative benchmarking metric, since top 3 may not reflect the real performance of the model. Instead, we encourage reporting the statistics of the optimized distribution. Figure \ref{fig:qed} shows that MolecularRNN shifts the distribution father to the maximum values of QED compared to GCPN.



\paragraph{Melting temperature maximization.} We train a graph convolution regression model introduced in \citep{kipf2016graphconv} for predicting the melting point of a molecule. Training and test datasets were $37940$ and $9458$ objects correspondingly; with $T_{melt}$ ranging from $-196^{\circ}C$ to $517^{\circ}C$. The model has $4$ layers with hidden sizes of $128$. We use Adam optimizer, starting with a learning rate of $0.001$ and exponential decay with $\gamma=0.8$ after every epoch. The model is trained with a batch size of 32 for 30 epochs. The model converges to RMS error of $39.5^{\circ}C$,  that is comparable to the state-of-the-art for the same dataset \citep{tetko2014}. This model is then used to assign a reward function
$r(mol) = \exp(t_{pred}(mol) + 1)$,
where $t_{pred}(mol)$ is the normalized predicted melting temperature for a molecule. 

\begin{figure}[h!]
\centering
\subfloat[Distribution of predicted melting temperature for base and optimized models]{\includegraphics[width=0.44\linewidth]{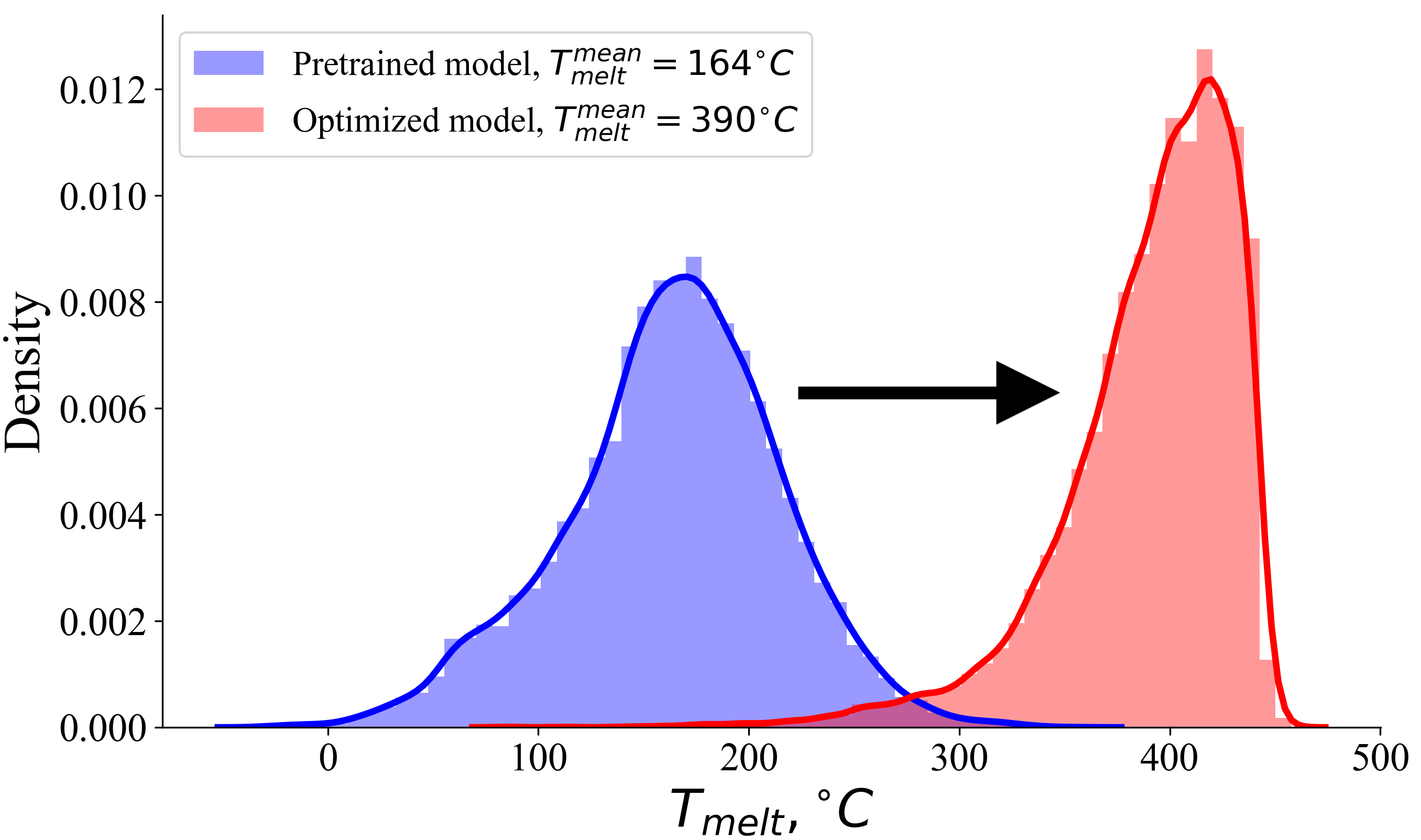}\label{fig:melt_temp_dist}}
\qquad
\subfloat[Examples of generated molecules with highest melting temperature]{\includegraphics[width=0.44\linewidth]{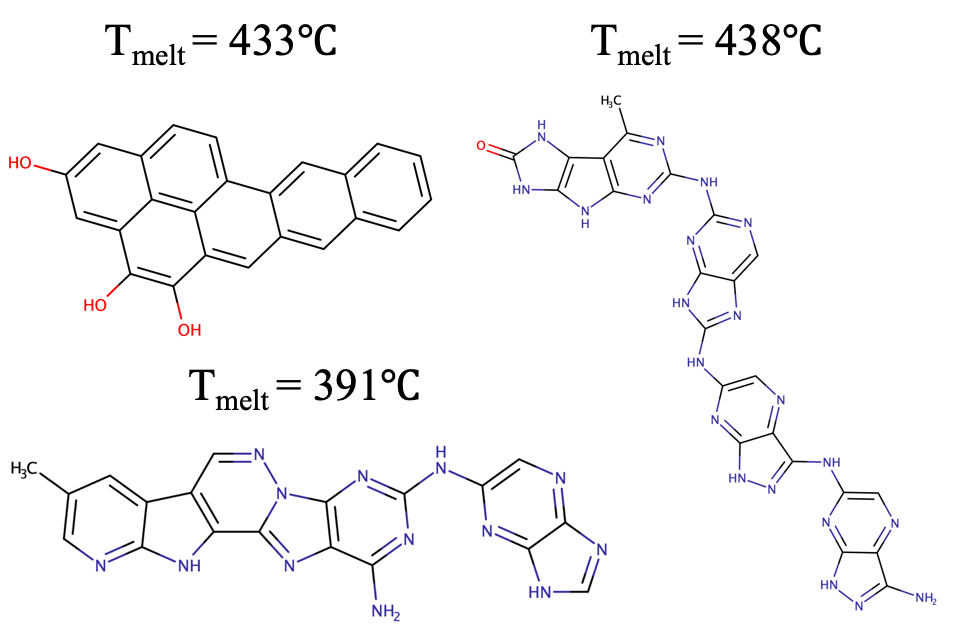}\label{fig:melt_temp_examples}}
\caption{Melting temperature maximization}
\label{fig:metl_temp}
\end{figure}

For this experiment, we used model pretrained on ChEMBL dataset and optimized it with the same settings as in the previous experiments -- 300 iterations with a batch size of 512 and Adam optimizer with a constant learning rate of $10^{-5}$. Figure~\ref{fig:melt_temp_dist} shows the relative distribution shift of predicted property for the molecules sampled from the pretrained model and for the molecules sampled from the optimized model. Example of generated molecules with predicted values of $T_{melt}$ are shown in Figure~\ref{fig:melt_temp_examples}. Interestingly, in this experiment, MolecularRNN rediscovered two known chemical phenomena. First, fusing multiple aromatic rings significantly increases the $T_{melt}$. Second, the presence of C=O, OH, $NH_{2}$ and heterocyclic nitrogens make molecules more polar. This usually enhances dipole-dipole interactions and subsequently increase $T_{melt}$ as well.

\section{Summary}

We proposed MolecularRNN, the model for generating realistic molecular graphs. MolecularRNN learns diverse distributions through unsupervised pretraining, generating 100\% valid molecules in inference, while still receiving negative feedback from invalid ones during training. Combined with policy gradient optimization, MolecularRNN solves the problem of generating molecules with desired properties. 
Optimized MolecularRNN outperforms other state-of-the-art methods on the benchmark tasks. Furthermore, we use the predictive model as a critic to optimize melting temperature, a property that cannot be calculated from a molecular graph.
Further studies address problems of multi-objective property optimization and completion of a molecular graph from a given scaffold. 

\section*{Acknowledgments}
O.I. acknowledges support from DOD-ONR (N00014-16-1-2311), National Science Foundation (NSF CHE-1802789), and Eshelman Institute for Innovation (EII) awards. M.P. acknowledges The Molecular Sciences Software Institute (MolSSI) Software Fellowship and NVIDIA Graduate Fellowship. We gratefully acknowledge the support and hardware donation from NVIDIA Corporation.

\medskip
 
\bibliography{references}

\begin{thebibliography}{24}
\providecommand{\natexlab}[1]{#1}
\providecommand{\url}[1]{\texttt{#1}}
\expandafter\ifx\csname urlstyle\endcsname\relax
  \providecommand{\doi}[1]{doi: #1}\else
  \providecommand{\doi}{doi: \begingroup \urlstyle{rm}\Url}\fi

\bibitem[Schneider and Fechner(2005)]{schneider2005computer}
Gisbert Schneider and Uli Fechner.
\newblock Computer-based de novo design of drug-like molecules.
\newblock \emph{Nature Reviews Drug Discovery}, 4\penalty0 (8):\penalty0 649,
  2005.

\bibitem[Butler et~al.(2018)Butler, Davies, Cartwright, Isayev, and
  Walsh]{butler2018ml}
Keith~T Butler, Daniel~W Davies, Hugh Cartwright, Olexandr Isayev, and Aron
  Walsh.
\newblock Machine learning for molecular and materials science.
\newblock \emph{Nature}, 559\penalty0 (7715):\penalty0 547, 2018.

\bibitem[Weininger(1988)]{weininger1988}
David Weininger.
\newblock Smiles, a chemical language and information system. 1. introduction
  to methodology and encoding rules.
\newblock \emph{Journal of chemical information and computer sciences},
  28\penalty0 (1):\penalty0 31--36, 1988.

\bibitem[Olivecrona et~al.(2017)Olivecrona, Blaschke, Engkvist, and
  Chen]{olivecrona2017molecular}
Marcus Olivecrona, Thomas Blaschke, Ola Engkvist, and Hongming Chen.
\newblock Molecular de-novo design through deep reinforcement learning.
\newblock \emph{Journal of cheminformatics}, 9\penalty0 (1):\penalty0 48, 2017.

\bibitem[Gómez-Bombarelli et~al.(2018)Gómez-Bombarelli, Wei, Duvenaud,
  Hernández-Lobato, Sánchez-Lengeling, Sheberla, Aguilera-Iparraguirre,
  Hirzel, Adams, and Aspuru-Guzik]{GomezBombarelli}
Rafael Gómez-Bombarelli, Jennifer~N. Wei, David Duvenaud, José~Miguel
  Hernández-Lobato, Benjamín Sánchez-Lengeling, Dennis Sheberla, Jorge
  Aguilera-Iparraguirre, Timothy~D. Hirzel, Ryan~P. Adams, and Alán
  Aspuru-Guzik.
\newblock Automatic chemical design using a data-driven continuous
  representation of molecules.
\newblock \emph{ACS Central Science}, 4\penalty0 (2):\penalty0 268--276, 2018.
\newblock \doi{10.1021/acscentsci.7b00572}.
\newblock URL \url{https://doi.org/10.1021/acscentsci.7b00572}.

\bibitem[Popova et~al.(2018)Popova, Isayev, and Tropsha]{popova2018deep}
Mariya Popova, Olexandr Isayev, and Alexander Tropsha.
\newblock Deep reinforcement learning for de novo drug design.
\newblock \emph{Science advances}, 4\penalty0 (7):\penalty0 eaap7885, 2018.

\bibitem[Jin et~al.(2018)Jin, Barzilay, and Jaakkola]{jin2018junctiontree}
Wengong Jin, Regina Barzilay, and Tommi Jaakkola.
\newblock Junction tree variational autoencoder for molecular graph generation.
\newblock \emph{arXiv preprint arXiv:1802.04364}, 2018.

\bibitem[Li et~al.(2018{\natexlab{a}})Li, Vinyals, Dyer, Pascanu, and
  Battaglia]{li2018learning}
Yujia Li, Oriol Vinyals, Chris Dyer, Razvan Pascanu, and Peter Battaglia.
\newblock Learning deep generative models of graphs.
\newblock \emph{arXiv preprint arXiv:1803.03324}, 2018{\natexlab{a}}.

\bibitem[You et~al.(2018{\natexlab{a}})You, Liu, Ying, Pande, and
  Leskovec]{you2018graphconvpolicy}
Jiaxuan You, Bowen Liu, Zhitao Ying, Vijay Pande, and Jure Leskovec.
\newblock Graph convolutional policy network for goal-directed molecular graph
  generation.
\newblock In \emph{Advances in Neural Information Processing Systems}, pages
  6410--6421, 2018{\natexlab{a}}.

\bibitem[Liu et~al.(2018)Liu, Allamanis, Brockschmidt, and
  Gaunt]{liu2018constrainedgraph}
Qi~Liu, Miltiadis Allamanis, Marc Brockschmidt, and Alexander Gaunt.
\newblock Constrained graph variational autoencoders for molecule design.
\newblock In \emph{Advances in Neural Information Processing Systems}, pages
  7795--7804, 2018.

\bibitem[Li et~al.(2018{\natexlab{b}})Li, Zhang, and Liu]{li2018multi}
Yibo Li, Liangren Zhang, and Zhenming Liu.
\newblock Multi-objective de novo drug design with conditional graph generative
  model.
\newblock \emph{Journal of cheminformatics}, 10\penalty0 (1):\penalty0 33,
  2018{\natexlab{b}}.

\bibitem[You et~al.(2018{\natexlab{b}})You, Ying, Ren, Hamilton, and
  Leskovec]{you2018graphrnn}
Jiaxuan You, Rex Ying, Xiang Ren, William~L Hamilton, and Jure Leskovec.
\newblock Graphrnn: Generating realistic graphs with deep auto-regressive
  models.
\newblock \emph{arXiv preprint arXiv:1802.08773}, 2018{\natexlab{b}}.

\bibitem[Segler et~al.(2017)Segler, Kogej, Tyrchan, and
  Waller]{segler2017generating}
Marwin~HS Segler, Thierry Kogej, Christian Tyrchan, and Mark~P Waller.
\newblock Generating focused molecule libraries for drug discovery with
  recurrent neural networks.
\newblock \emph{ACS central science}, 4\penalty0 (1):\penalty0 120--131, 2017.

\bibitem[Kadurin et~al.(2017)Kadurin, Nikolenko, Khrabrov, Aliper, and
  Zhavoronkov]{kadurin2017drugan}
Artur Kadurin, Sergey Nikolenko, Kuzma Khrabrov, Alex Aliper, and Alex
  Zhavoronkov.
\newblock drugan: an advanced generative adversarial autoencoder model for de
  novo generation of new molecules with desired molecular properties in silico.
\newblock \emph{Molecular pharmaceutics}, 14\penalty0 (9):\penalty0 3098--3104,
  2017.

\bibitem[Landrum et~al.(2006)]{landrum2006rdkit}
Greg Landrum et~al.
\newblock Rdkit: Open-source cheminformatics, 2006.

\bibitem[Williams(1987)]{williams1987}
R.~Williams.
\newblock A class of gradient-estimation algorithms for reinforcement learning
  in neural networks.
\newblock In \emph{International Conference on Neural Networks}, 1987.

\bibitem[Ertl and Schuffenhauer(2009)]{ertl2009estimation}
Peter Ertl and Ansgar Schuffenhauer.
\newblock Estimation of synthetic accessibility score of drug-like molecules
  based on molecular complexity and fragment contributions.
\newblock \emph{Journal of cheminformatics}, 1\penalty0 (1):\penalty0 8, 2009.

\bibitem[Bickerton et~al.(2012)Bickerton, Paolini, Besnard, Muresan, and
  Hopkins]{bickerton2012quantifying}
G~Richard Bickerton, Gaia~V Paolini, J{\'e}r{\'e}my Besnard, Sorel Muresan, and
  Andrew~L Hopkins.
\newblock Quantifying the chemical beauty of drugs.
\newblock \emph{Nature chemistry}, 4\penalty0 (2):\penalty0 90, 2012.

\bibitem[Polykovskiy et~al.(2018)Polykovskiy, Zhebrak, Sanchez-Lengeling,
  Golovanov, Tatanov, Belyaev, Kurbanov, Artamonov, Aladinskiy, Veselov,
  et~al.]{polykovskiy2018molecular}
Daniil Polykovskiy, Alexander Zhebrak, Benjamin Sanchez-Lengeling, Sergey
  Golovanov, Oktai Tatanov, Stanislav Belyaev, Rauf Kurbanov, Aleksey
  Artamonov, Vladimir Aladinskiy, Mark Veselov, et~al.
\newblock Molecular sets (moses): A benchmarking platform for molecular
  generation models.
\newblock \emph{arXiv preprint arXiv:1811.12823}, 2018.

\bibitem[Gaulton et~al.(2011)Gaulton, Bellis, Bento, Chambers, Davies, Hersey,
  Light, McGlinchey, Michalovich, Al-Lazikani, and Overington]{gaulton2011}
A.~Gaulton, L.J. Bellis, A.P. Bento, J.~Chambers, M.~Davies, A.~Hersey,
  Y.~Light, S.~McGlinchey, D.~Michalovich, B.~Al-Lazikani, and J.P. Overington.
\newblock Chembl: a large-scale bioactivity database for drug discovery.
\newblock \emph{Nucleic acids research}, 40\penalty0 (D1):\penalty0
  D1100--D1107, 2011.

\bibitem[Irwin and Shoichet(2005)]{irwin2005zinc}
John~J Irwin and Brian~K Shoichet.
\newblock Zinc - a free database of commercially available compounds for
  virtual screening.
\newblock \emph{Journal of chemical information and modeling}, 45\penalty0
  (1):\penalty0 177--182, 2005.

\bibitem[Guimaraes et~al.(2017)Guimaraes, Sanchez-Lengeling, Outeiral, Farias,
  and Aspuru-Guzik]{guimaraes2017objective}
Gabriel~Lima Guimaraes, Benjamin Sanchez-Lengeling, Carlos Outeiral, Pedro
  Luis~Cunha Farias, and Al{\'a}n Aspuru-Guzik.
\newblock Objective-reinforced generative adversarial networks (organ) for
  sequence generation models.
\newblock \emph{arXiv preprint arXiv:1705.10843}, 2017.

\bibitem[Kipf and Welling(2016)]{kipf2016graphconv}
Thomas~N Kipf and Max Welling.
\newblock Semi-supervised classification with graph convolutional networks.
\newblock \emph{arXiv preprint arXiv:1609.02907}, 2016.

\bibitem[Tetko et~al.(2014)Tetko, Novotarskyi, Patiny, Kondratov, Petrenko,
  Charochkina, and Asiri]{tetko2014}
Y.~Tetko, I.V. ans~Sushko, S.~Novotarskyi, L.~Patiny, I.~Kondratov, A.E.
  Petrenko, L.~Charochkina, and A.M. Asiri.
\newblock How accurately can we predict the melting points of drug-like
  compounds?
\newblock \emph{Journal of chemical information and modeling}, 54\penalty0
  (12):\penalty0 D1100--D1107, 2014.

\end{thebibliography}

\end{document}